\documentclass[11pt]{article}

\usepackage[a4paper,margin=1in]{geometry}
\usepackage[numbers]{natbib}
\usepackage{amsmath,amssymb}
\usepackage{graphicx}
\usepackage{booktabs}
\usepackage{hyperref}
\usepackage{placeins}
\usepackage{subcaption}
\usepackage{multirow}
\usepackage[final]{microtype}
\usepackage{caption}
\usepackage{authblk}
\usepackage{enumitem}

\hypersetup{hidelinks}
\title{SPECTRA-Net: Scalable Pipeline for Explainable Cross-domain Tensor Representations for AI-generated Images Detection}

\author[1]{Sarra Arab}
\author[1]{Anfal Achouri}
\author[1,*]{Seif Eddine Bouziane}
\affil[1]{The National School of Artificial Intelligence, Algiers, Algeria}
\affil[*]{Corresponding author: \texttt{seifeddine.bouziane@ensia.edu.dz}}

\date{}

\begin{document}

\maketitle

\begin{abstract}
The rapid proliferation of AI-generated images (AIGI) presents a significant challenge to
digital information integrity. While human observers and existing detection models
struggle to keep pace with the increasing sophistication of generative models, the
need for robust, real-time detection systems has become critical. This paper
introduces SPECTRA-Net, a scalable pipeline for explainable, cross-domain tensor
representations for AIGI detection. Our approach leverages a multi-view
representation of images, combining global semantic features from a Vision
Foundation Model (VFM), spectral analysis, local patch-based anomaly detection, and
statistical descriptors. By fusing these complementary data streams, SPECTRA-Net
achieves state-of-the-art performance in both in-domain and cross-domain settings,
demonstrating high accuracy and generalization capabilities across a wide range of
challenging datasets, including WildFake, Chameleon, and RRDataset. The proposed
pipeline not only provides a robust solution for AIGI detection but also offers
explainability through artifact localization, paving the way for more trustworthy and
reliable content verification in real-world applications.
\end{abstract}


\noindent\textbf{Main contributions.}
\begin{itemize}
\item Multi-view framework integrating semantic, frequency-domain, statistical, and patch-level representations.
\item Robust detection pipeline resilient to real-world degradations, including JPEG compression and Gaussian blur.
\item Patch-wise explainability enabling localized artifact detection and spatial interpretability.
\end{itemize}

\section{Introduction}
Recent advances in generative models have led to an incredible growth in highly realistic AI-generated images. Modern diffusion models \cite{higham2023diffusion} and large-scale generative frameworks are now capable of producing visual content that is often indistinguishable from real photographs, accelerating their adoption across industries and social media platforms; however, this rapid progress also raises serious concerns regarding image authenticity and misinformation, motivating the need for reliable AI-generated image detection systems.

Early detection approaches primarily focused on identifying low-level artifacts introduced by GAN-based generators, such as color inconsistencies, checkerboard patterns, or spectral irregularities \cite{frank2020leveraging}. These artifact-driven methods are mainly effective against early-generation models but struggle with diffusion-based synthesis, which relies on iterative denoising processes that leave fewer explicit traces \cite{dhariwal2021diffusion}, leading to significant performance degradation when detectors trained under GAN assumptions are exposed to modern generative models.

More recent learning-based approaches leverage deep convolutional networks, transformers, or vision foundation models to learn discriminative representations directly from image data. Foundation-model-based detectors, such as DINO-Detect, demonstrate strong robustness to degradations like motion blur by distilling semantic knowledge from frozen DINO backbones \cite{shen2025dinodetect}. However, such methods primarily emphasize high-level semantics and may overlook complementary low-level cues that remain informative under real-world distortions. Similarly, lightweight and real-time architectures, including SAPN \cite{gangarapu2025sapn} and binary neural network–based detectors \cite{fasterthanlies2024}, prioritize efficiency and deployability but often exhibit limited cross-dataset generalization or sensitivity to domain shifts.

Another line of work explores frequency-domain and texture-based representations, motivated by the observation that generative models struggle to reproduce natural spectral statistics. Frequency-aware detectors \cite{durall2020watch} and hybrid RGB–FFT pipelines \cite{fasterthanlies2024} have shown promising results, particularly under controlled settings. Recent advances in dual-domain processing, such as wavelet-based diffusion models for blind image separation \cite{gong2026dbidm}, demonstrate the effectiveness of combining frequency and spatial features for artifact removal. Complementarily, patch-based and pixel-level approaches such as LaDeDa \cite{cavia2024realtimedeepfakedetectionrealworld} and PatchCraft \cite{patchcraft2023} highlight the importance of localized texture inconsistencies and enable spatial explainability through anomaly heatmaps. While effective, these methods often rely on a single representation domain—either frequency, spatial texture, or semantic features—which limits robustness under diverse real-world transformations such as JPEG compression, Gaussian blur, and platform re-encoding, where single-modality detectors may fail when their specific cues are degraded. Recent multi-level feature representation approaches~\cite{XU202376} have shown promise by combining CNNs with vision transformers for enhanced forensic analysis, yet the integration of global semantic knowledge with complementary spectral and patch-level cues remains underexplored.

To address these challenges, we introduce SPECTRA-Net, a multi-view detection framework designed for robust and explainable AIGI detection under realistic, in-the-wild conditions. Rather than relying on a single cue, SPECTRA-Net integrates complementary semantic, frequency-domain, statistical, and patch-level representations to improve generalization and resilience to real-world degradations. By combining global semantic understanding with spectral analysis and localized anomaly detection, the proposed framework bridges the gap between accuracy, robustness, and interpretability.

\section{Methodology}
\subsection{Training Dataset}
\label{sec:dataset_wildfake}
We leverage WildFake~\cite{liu2023wildfake}, a large-scale dataset designed for AI-generated image detection in wild, real-world conditions rather than controlled, lab-curated settings. Reflecting the diversity and variability of images shared on social media platforms, WildFake contains over 3.7 million images generated by state-of-the-art models, including GANs and diffusion-based systems such as Stable Diffusion and Midjourney. For our baseline SPECTRA-Net training, we focus on diffusion-based subsets, excluding GAN content to target contemporary generative artifacts. Its scale and natural, in-the-wild distribution enable the model to learn robust and generalizable representations beyond idealized laboratory benchmarks.
\subsection{Finetuning Dataset}
\label{sec:dataset_rrdataset}
We use RRDataset~\cite{li2025bridginggapidealrealworld}, containing 10,000 real and 10,000 AI-generated images across seven scenarios (six high-risk + everyday life), all subjected to 2--6 rounds of internet transmission across eight platforms and four re-digitization processes (scanning, printing, photographing, screen capture).  

The baseline SPECTRA-Net was first evaluated on all splits (train, validation, test) to establish reference performance. We then fine-tuned it on the training and validation splits, testing the fine-tuned model exclusively on the RRDataset test split (original, re-digitized, and transfer variants) to measure generalization to unseen, challenging content.

\subsection{Overview}
Our proposed framework, SPECTRA-Net, moves beyond conventional binary detection of lab-curated synthetic image media by addressing the more challenging problem of distinguishing AI-generated images from real content curated on social media platforms. The framework is explicitly designed to operate under realistic in-the-wild degradations, including severe JPEG compression and blur, thereby closely mimicking the conditions encountered in real-world deployments while maintaining robust detection performance.

Beyond robustness, SPECTRA-Net delivers explainable predictions through a patch-wise scoring strategy that produces interpretable spatial heatmaps, revealing the regions that most strongly influence the final decision.

To jointly meet the requirements of robustness, explainability, and efficient inference across diverse visual domains, SPECTRA-Net adopts a multi-stage pipeline comprising two principal components: (i) an offline pre-processing unit for constructing multi-view image representations, and (ii) a training unit dedicated to learning discriminative generative patterns.

\begin{figure}[!htbp]
    \centering
    \includegraphics[width=\textwidth,height=0.80\textheight,keepaspectratio]{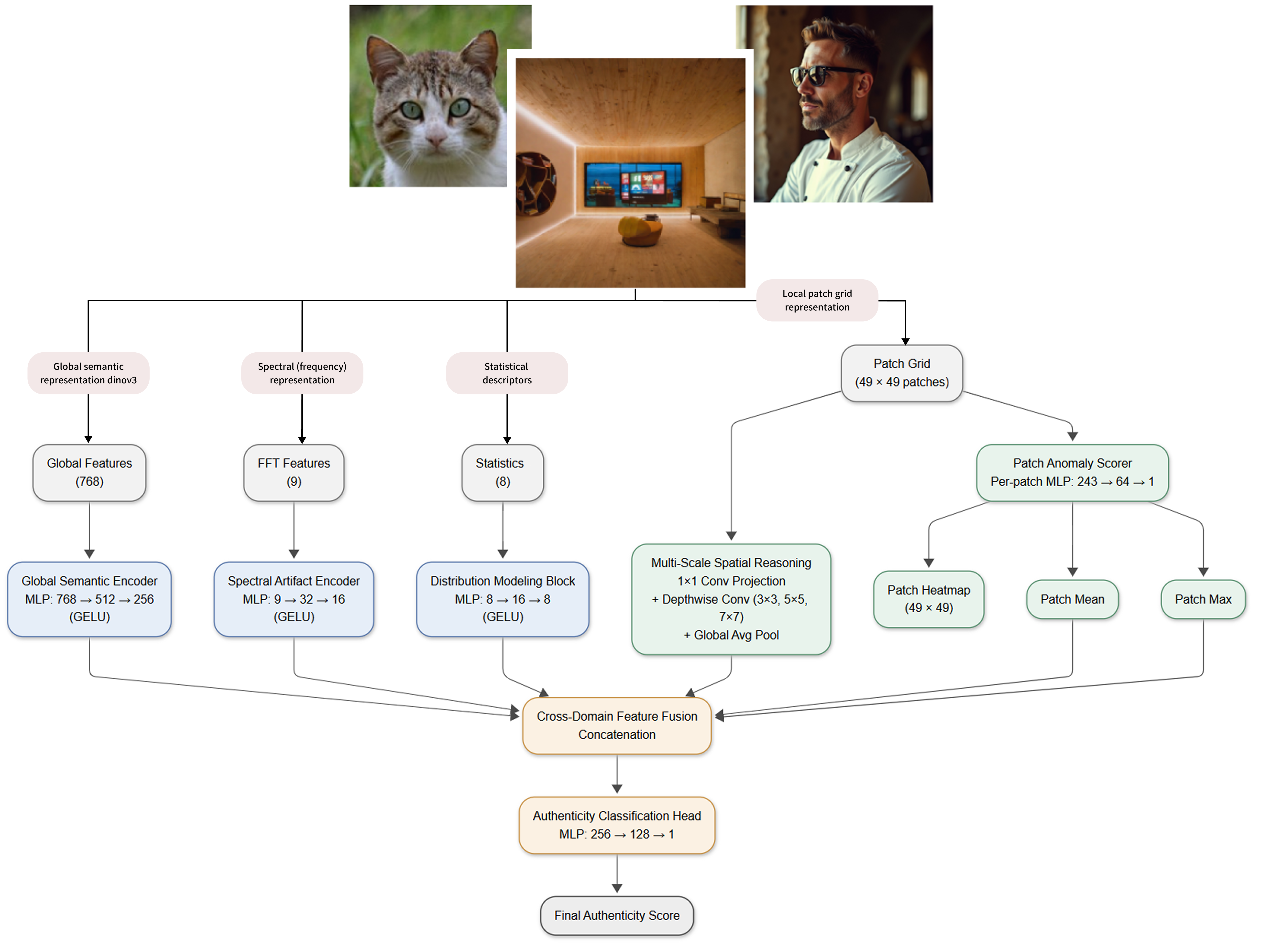}
    \caption{SPECTRA-Net architecture with patch-wise explainability and cross-domain feature fusion.}
    \label{fig:spectranet_arch}
\end{figure}

As illustrated in Fig.~\ref{fig:spectranet_arch}, the pre-processing unit transforms raw input images into a standardized multi-view tensor representation. This representation is subsequently utilized by the training unit to learn discriminative cues that differentiate real images from AI-generated content, enabling efficient and interpretable inference at deployment time.

\subsection{Pre-processing Stage: Multi-view Representation}
\label{sec:preprocessing}

Given an input RGB image $\mathbf{I} \in \mathbb{R}^{H_0 \times W_0 \times 3}$, we construct a comprehensive multi-view representation that jointly captures semantic, frequency-domain, statistical, and spatial characteristics. The representation combines four complementary feature types: (1) global semantic descriptors (768D), (2) frequency-domain statistics (9D), (3) image statistics (8D), and (4) local spatial patches (49×243D). This combined representation enables robust detection of diffusion-model-generated artifacts across multiple levels of abstraction.

\subsubsection{Image Normalization and Preprocessing}

Each image is first resized to a fixed resolution of $H = W = 448$ pixels and normalized using ImageNet statistics ~\cite{deng2009imagenet,dosovitskiy2021vit}:
\begin{equation}
\mathbf{X} = \frac{\mathbf{I}/255 - \boldsymbol{\mu}}{\boldsymbol{\sigma}},
\label{eq:norm}
\end{equation}
where $\boldsymbol{\mu} = [0.485, 0.456, 0.406]$ and $\boldsymbol{\sigma} = [0.229, 0.224, 0.225]$ are per-channel ImageNet mean and standard deviation. The normalized tensor is denoted $\mathbf{X} \in \mathbb{R}^{448 \times 448 \times 3}$.

\subsubsection{Feature Extraction: Global Semantic Representation (768D)}

High-level semantic features are extracted using a DINOv3 Vision Transformer backbone (ViT-Base-16)~\cite{simeoni2025dinov3,dosovitskiy2021vit}, which encodes the normalized image as:
\begin{equation}
\mathbf{F} = \text{ViT}(\mathbf{X}) \in \mathbb{R}^{(1+N) \times 768},
\label{eq:vit}
\end{equation}
where $N = (448/16)^2 = 784$ denotes the number of patch tokens and $768$ is the embedding dimension of the base model. We retain only the class token, which serves as a compact global image descriptor:
\begin{equation}
\mathbf{f}_{\text{global}} = \mathbf{F}[0, :] \in \mathbb{R}^{768}.
\label{eq:cls}
\end{equation}

\subsubsection{Feature Extraction: Frequency-Domain Representation (9D)}

Diffusion-generated images exhibit characteristic high-frequency patterns distinct from natural images~\cite{durall2020watch,frank2020leveraging}. We extract frequency-domain statistics via 2D Fast Fourier Transform (FFT), a standard tool for analyzing spectral artifacts in synthetic imagery~\cite{gonzalez2008digital}. For each RGB channel $c \in \{R, G, B\}$:
\begin{equation}
\begin{aligned}
\mathcal{F}_c(u,v)
&= \sum_{x=0}^{H-1} \sum_{y=0}^{W-1} \mathbf{X}_c(x,y) \\
&\quad\cdot \exp\left(-j2\pi\left(\frac{ux}{H} + \frac{vy}{W}\right)\right),
\end{aligned}
\label{eq:fft}
\end{equation}

The magnitude and phase representations are computed as:
\begin{equation}
\mathbf{M}_c = |\mathcal{F}_c|, \quad \boldsymbol{\Phi}_c = \angle\mathcal{F}_c,
\label{eq:magphase}
\end{equation}

Log-compressed magnitudes emphasize high-frequency components characteristic of diffusion artifacts~\cite{mccloskey2018detectinggangeneratedimageryusing}:
\begin{equation}
\tilde{\mathbf{M}}_c = \log(1 + \mathbf{M}_c),
\label{eq:logmag}
\end{equation}

For each channel, spatial-domain magnitude statistics are extracted:
\begin{equation}
\mu_c = \mathbb{E}[\tilde{\mathbf{M}}_c], \quad \sigma_c = \sqrt{\mathbb{E}[(\tilde{\mathbf{M}}_c - \mu_c)^2]},
\label{eq:magstats}
\end{equation}

Phase entropy measures the randomness and structural coherence of phase information~\cite{wang2020cnn}:
\begin{equation}
\eta_c = \sqrt{\mathbb{E}\left[\left(\frac{\boldsymbol{\Phi}_c + \pi}{2\pi} - 0.5\right)^2\right]},
\label{eq:phaseentropy}
\end{equation}

The complete frequency-domain feature vector combines these statistics across three channels:
\begin{equation}
\mathbf{f}_{\text{fft}} = [\mu_R, \sigma_R, \eta_R, \mu_G, \sigma_G, \eta_G, \mu_B, \sigma_B, \eta_B]^T \in \mathbb{R}^9,
\label{eq:fft_features}
\end{equation}

\subsubsection{Feature Extraction: Statistical Descriptors (8D)}

To capture spatial-domain intensity statistics complementary to the frequency-domain representation, we compute eight descriptors from the normalized image channels~\cite{haralick1973textural}. Let $\mathbf{X}_c \in \mathbb{R}^{H \times W}$ be the pixel intensities of channel $c \in \{R, G, B\}$ in $\mathbf{X}$. The statistical feature vector is then:
\begin{equation}
\mathbf{f}_{\text{stat}} =
[\mu'_R, \sigma'_R, \mu'_G, \sigma'_G, \mu'_B, \sigma'_B, \gamma, \kappa]^T
\in \mathbb{R}^8,
\label{eq:stat_features}
\end{equation}

where the per-channel mean and standard deviation are given by:
\begin{equation}
\mu'_c = \mathbb{E}[\mathbf{X}_c], \quad
\sigma'_c = \sqrt{\mathbb{E}\big[(\mathbf{X}_c - \mu'_c)^2\big]},
\label{eq:channel_stats}
\end{equation}

and the global skewness and excess kurtosis, computed over all spatial locations and channels, are defined as:
\begin{equation}
\gamma = \frac{\mathbb{E}\big[(\mathbf{X} - \mu')^3\big]}{(\sigma')^3}, \quad
\kappa = \frac{\mathbb{E}\big[(\mathbf{X} - \mu')^4\big]}{(\sigma')^4} - 3,
\label{eq:higher_moments}
\end{equation}
with $\mu'$ and $\sigma'$ denoting the global mean and standard deviation of $\mathbf{X}$.

\subsubsection{Feature Extraction: Local Spatial Representation (49×243D)}

Fine-grained spatial information is preserved by extracting non-overlapping $9 \times 9$ patches from the normalized image. Using 2D unfold operation with kernel size $9 \times 9$ and stride $9$:
\begin{equation}
\mathbf{P} = \text{unfold}(\mathbf{X}, \text{kernel}=9 \times 9, \text{stride}=9) \in \mathbb{R}^{2401 \times 243},
\label{eq:patches}
\end{equation}
For an input image with $H=W=448$, using kernel size and stride $9$ yields 
$N_p=\left(\lfloor\frac{H-9}{9}\rfloor+1\right)^2=49^2=2401$ non-overlapping patches arranged on a $49\times49$ grid. 
Each $9\times9\times3$ patch is vectorized into a $243$-dimensional feature, producing 
$\mathbf{P}\in\mathbb{R}^{2401\times243}$ via the unfold operation, which can be reshaped into a structured $49\times49$ grid for spatial reasoning. 
Such patch-based representations preserve local structure~\cite{phan2022patch} and provide the foundation for patch-wise anomaly detection.

\subsubsection{Multi-view Representation Summary}

The complete multi-view representation is:
\begin{equation}
\mathcal{R}_{\text{pre}} = \{\mathbf{f}_{\text{global}}, \mathbf{f}_{\text{fft}}, \mathbf{f}_{\text{stat}}, \mathbf{P}\} \in \mathbb{R}^{768} \times \mathbb{R}^9 \times \mathbb{R}^8 \times \mathbb{R}^{2401 \times 243},
\label{eq:multiview_preproc}
\end{equation}

\subsection{Training Stage: Neural Feature Encoding and Classification}
\label{sec:training}

The training stage receives the multi-view representation $\mathcal{R}_{\text{pre}}$ \eqref{eq:multiview_preproc} from the pre-processing stage and learns to encode these features through specialized neural network modules with GELU~\cite{hendrycks2016gaussian} activation functions.

\subsubsection{Global Semantic Feature Encoding}

The 768-dimensional global semantic features $\mathbf{f}_{\text{global}}$ \eqref{eq:cls} are encoded through a Global Semantic Encoder (MLP: $768 \to 512 \to 256$) to produce task-aware semantic embeddings:
\begin{equation}
\mathbf{f}'_{\text{global}} = \mathbf{W}_2 \cdot \text{GELU}(\mathbf{W}_1 \mathbf{f}_{\text{global}} + \mathbf{b}_1) + \mathbf{b}_2 \in \mathbb{R}^{256}
\label{eq:global_encoding}
\end{equation}
where intermediate dimension is 512, with Dropout(0.3) applied between layers.

\subsubsection{Spectral Feature Encoding}

The 9-dimensional frequency-domain features $\mathbf{f}_{\text{fft}}$ \eqref{eq:fft_features} are encoded through a Spectral Artifact Encoder (MLP: $9 \to 32 \to 16$) that expands then compresses to capture frequency-space anomalies:
\begin{equation}
\mathbf{f}'_{\text{fft}} = \mathbf{W}'_2 \cdot \text{GELU}(\mathbf{W}'_1 \mathbf{f}_{\text{fft}} + \mathbf{b}'_1) + \mathbf{b}'_2 \in \mathbb{R}^{16}
\label{eq:fft_encoding}
\end{equation}
where the intermediate expansion to 32 dimensions allows cross-channel interactions before compression to 16 dimensions.

\subsubsection{Statistical Feature Encoding}

The 8-dimensional statistical descriptors $\mathbf{f}_{\text{stat}}$ \eqref{eq:stat_features} are encoded through a Distribution Modeling Block (MLP: $8 \to 16 \to 8$):
\begin{equation}
\mathbf{f}'_{\text{stat}} = \mathbf{W}''_2 \cdot \text{GELU}(\mathbf{W}''_1 \mathbf{f}_{\text{stat}} + \mathbf{b}''_1) + \mathbf{b}''_2 \in \mathbb{R}^{8}
\label{eq:stat_encoding}
\end{equation}

\subsubsection{Spatial Patch Feature Encoding and Anomaly Scoring}

Each of the 2401 spatial patches is individually scored for anomalies. Each 243-dimensional patch is processed through a Per-patch Anomaly Scorer (MLP: $243 \to 64 \to 1$):
\begin{equation}
s_i = \mathbf{W}^{(p)}_2 \cdot 
\text{GELU}(\mathbf{W}^{(p)}_1 \mathbf{P}_i + \mathbf{b}^{(p)}_1) 
+ \mathbf{b}^{(p)}_2 \in \mathbb{R}, 
\quad i = 1, \ldots, 2401
\label{eq:patch_scoring}
\end{equation}
with Dropout(0.2) in the hidden layer. The scores $\{s_1, \ldots, s_{2401}\}$ can be reshaped into a $49 \times 49$ spatial heatmap for explainability. The multi-scale spatial reasoning block computes spatial correlations via depthwise convolutions with kernel sizes ($3 \times 3, 5 \times 5, 7 \times 7$).

\subsubsection{Cross-Domain Feature Fusion and Final Classification}

The encoded features from all four domains are concatenated to form a fused representation:
\begin{equation}
\mathbf{f}_{\text{fused}} = 
[\mathbf{f}'_{\text{global}};\ \mathbf{f}'_{\text{fft}};\ \mathbf{f}'_{\text{stat}};\ 
\mathbf{f}'_{\text{spatial}};\ \mathbf{p}_{\text{mean}};\ \mathbf{p}_{\text{max}}] 
\in \mathbb{R}^{D_{\text{fused}}}
\label{eq:fusion}
\end{equation}
This fused representation is passed through the final authenticity classification head (MLP: $D_{\text{fused}} \to 256 \to 128 \to 1$) which outputs a logit:
\begin{equation}
z = \mathbf{W}_{\text{clf},3} \cdot 
\text{GELU}\Big(
\mathbf{W}_{\text{clf},2} \cdot 
\text{GELU}(
\mathbf{W}_{\text{clf},1} \mathbf{f}_{\text{fused}} 
+ \mathbf{b}_{\text{clf},1}
)
+ \mathbf{b}_{\text{clf},2}
\Big)
+ \mathbf{b}_{\text{clf},3}
\label{eq:classification_logit}
\end{equation}

The predicted probability is computed via the sigmoid function:
\begin{equation}
\hat{y} = \sigma(z) = \frac{1}{1 + e^{-z}} \in [0, 1]
\label{eq:classification}
\end{equation}

\subsubsection{Optimization and Loss}

Training uses the AdamW optimizer~\cite{loshchilov2019decoupled} with learning rate $\eta = 2 \times 10^{-4}$, momentum parameters $\beta_1 = 0.9$ and $\beta_2 = 0.999$, and batch size 128. The loss function is Binary Cross-Entropy with Logits:
\begin{equation}
\mathcal{L} = -\frac{1}{B}\sum_{i=1}^{B} 
\left[
y_i \log(\sigma(z_i)) + (1 - y_i)\log(1 - \sigma(z_i))
\right]
\label{eq:loss}
\end{equation}
where $y_i \in \{0, 1\}$ is the ground truth label (0 = real, 1 = AI-generated) and $z_i$ is the predicted logit.

\subsubsection{Explainability via Patch-wise Heatmaps}

\begin{figure}[!htbp]
    \centering
    
    \begin{subfigure}[t]{0.48\columnwidth}
        \centering
        \includegraphics[width=\linewidth]{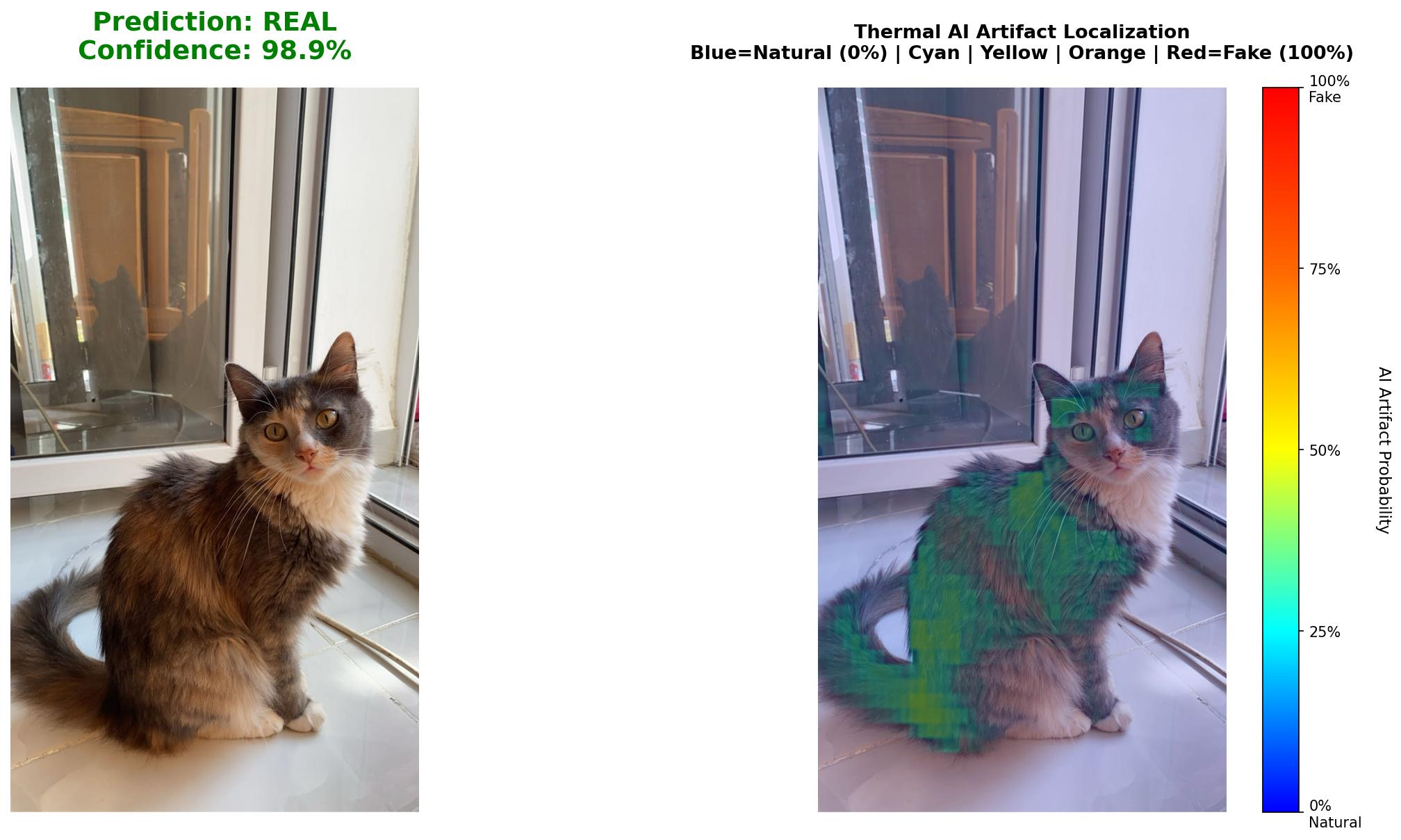}
        \caption{Ground-truth: Real}
        \label{fig:heatmap_real}
    \end{subfigure}
    \hfill
    \begin{subfigure}[t]{0.48\columnwidth}
        \centering
        \includegraphics[width=\linewidth]{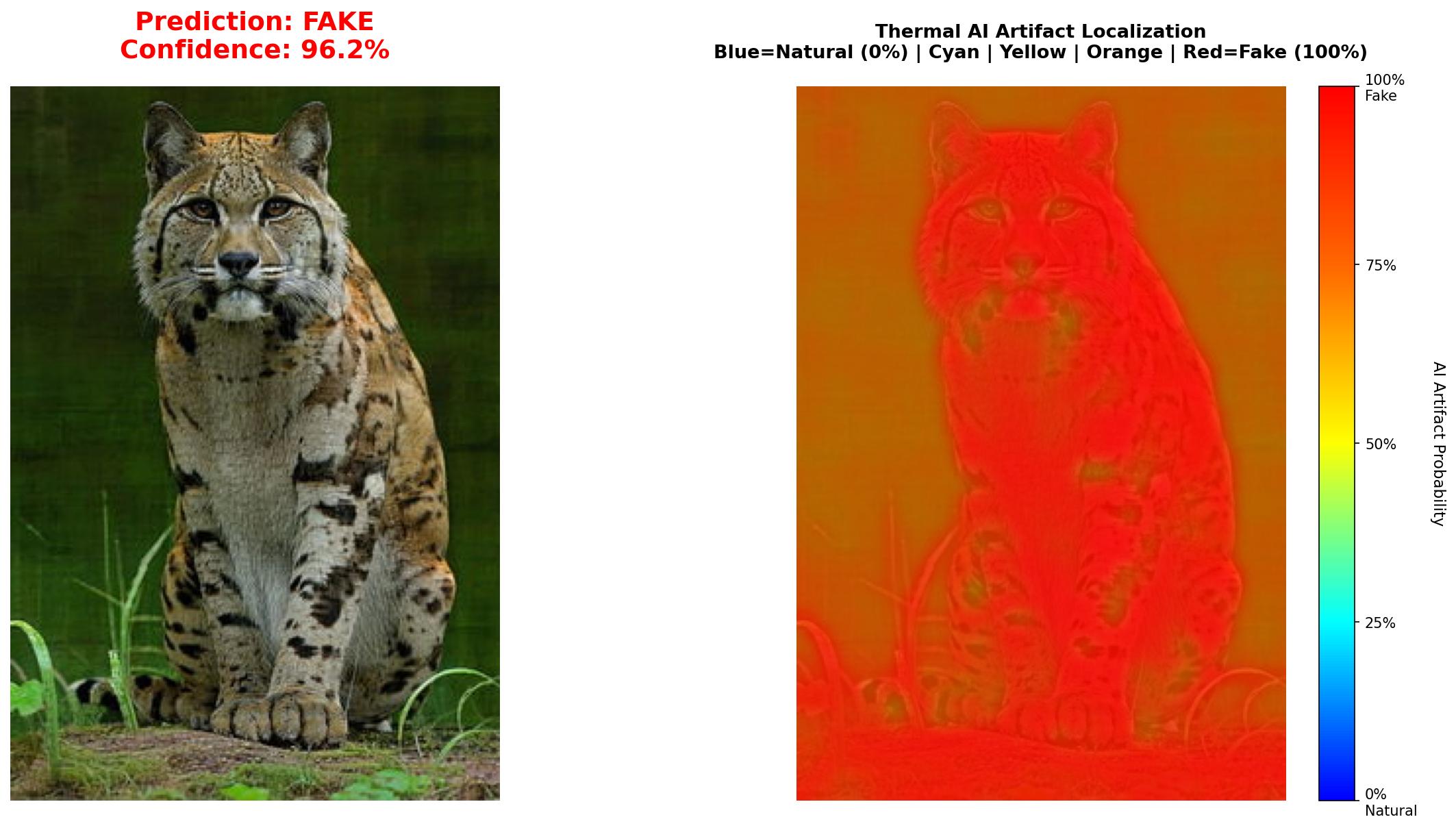}
        \caption{Ground-truth: Fake}
        \label{fig:heatmap_fake}
    \end{subfigure}
    
    \caption{Patch-wise explainability examples. Input images and corresponding $49\times49$ spatial heatmaps for (a) a real image and (b) an AI-generated image.}

    \label{fig:heatmap_combined}
\end{figure}

A key contribution of SPECTRA-Net is the generation of explainable predictions through patch-wise scoring. The $49 \times 49$ spatial heatmap can be visualized to show which image regions most strongly contribute to the final authenticity prediction, providing interpretability and localization of AI-generated artifacts (see Fig.~\ref{fig:heatmap_combined}).
\vspace{-20pt}
\section{Experimental Results}

\subsection{Evaluation Metrics}
\label{sec:metrics}

We evaluate detection performance using four complementary metrics to provide a comprehensive assessment of model robustness and generalization:

\subsubsection{Accuracy and AUC}

Accuracy (ACC) measures overall prediction correctness, while Area Under the Receiver Operating Characteristic Curve (AUC) provides threshold-independent discrimination assessment, particularly suitable for imbalanced datasets.

\subsubsection{Real Accuracy (Real-Acc) and Fake Accuracy (Fake-Acc)}

Real-Acc and Fake-Acc separately evaluate class-specific performance, essential for identifying asymmetric biases across real and fake classes.

\subsubsection{Mean Average Precision (mAP)} is computed exclusively for the WildRF benchmark to enable direct comparison with prior state-of-the-art methods:
\begin{equation}
\text{mAP} = \frac{1}{N} \sum_{i=1}^{N} \text{AP}_i,
\label{eq:map}
\end{equation}
where $\text{AP}_i$ is the average precision for class $i$ and $N$ is the number of classes (binary: $N=2$).

\subsection{Benchmark Datasets}
\label{sec:benchmarks}

Our evaluation encompasses four diverse benchmarks covering laboratory-controlled and real-world scenarios:

\subsubsection{Chameleon} Chameleon is a large-scale, manually curated benchmark containing approximately 26,000 test images spanning diverse object categories and scene compositions ~\cite{yan2025sanitycheckaigeneratedimage}. Designed as a sanity check for contemporary detection methods, Chameleon emphasizes deceptive realism and challenges models with high-quality, difficult-to-detect synthetic content across varied visual domains.

\subsubsection{WildRF} WildRF contains 5,351 images (train: 2,712 images; val: 398 images; test: 2,241 images) sourced from multiple internet platforms~\cite{cavia2024realtimedeepfakedetectionrealworld}, capturing real-world variability in image resolution, compression formats, post-processing edits, and generation techniques. This dataset reflects authentic distribution shifts encountered in practical forensic applications.

\subsubsection{Balanced Real vs. Recent AIGI Testing Dataset} 
From AI-GenBench~\cite{Pellegrini_2025}, which contains 180,000 images generated by 36 models spanning 2018–2024, we select 20,000 images from the five most recent generators: FLUX 1 Dev, FLUX 1 Schnell, DALL-E 3, Stable Diffusion XL 1.0, and Midjourney, capturing contemporary generative content.  

From BM-Real~\cite{bitmind_hf}, a collection of 28,393 high-resolution real images (192px–9,550px) on Hugging Face\footnote{\url{https://huggingface.co/datasets/bitmind/bm-real}}, we randomly sample 20,000 images to match the synthetic subset.  

Merging these two subsets, we obtain a 40,000-image testing dataset with equal representation of real and synthetic images, enabling robust evaluation of detection performance across diverse content.  

\subsubsection{RRDataset} RRDataset ~\cite{li2025bridginggapidealrealworld} contains 10,000 real + 10,000 synthetic images across seven scenarios (6 high-risk + everyday life). All content undergoes multi-round internet transmission (2--6 cycles across 8 platforms) and re-digitization such as scanning and printing. Test splits include original, re-digitized, and transfer variants, simulating practical deployment challenges.

\subsection{Baseline Results: SPECTRA-Net on Standard Benchmarks}
\label{sec:results_baseline}

Table~\ref{tab:chameleon_curated} presents the performance comparison on Chameleon and the balanced dataset.
\begin{table}[!htbp]
\centering
\scriptsize
\caption{Performance on Chameleon and Balanced Dataset.}
\label{tab:chameleon_curated}
\begin{tabular}{@{}llcccc@{}}
\hline
\textbf{Dataset} & \textbf{Method} & \textbf{ACC} & \textbf{AUC} & \textbf{Real Acc} & \textbf{Fake Acc} \\
 & & \textbf{(\%)} & & \textbf{(\%)} & \textbf{(\%)} \\
\hline
\multirow{2}{*}{Chameleon} 
& SOTA & 65.77 & -- & 95.06 & 25.8 \\
& \textbf{SN} & \textbf{85.70} & \textbf{0.946} & 82.51 & \textbf{89.95} \\
\hline
\multirow{2}{*}{Balanced} 
& SOTA & -- & -- & -- & -- \\
& \textbf{SN} & \textbf{92.11} & \textbf{0.975} & \textbf{95.2} & \textbf{89.0} \\
\hline
\end{tabular}

\end{table}

\noindent \textbf{SN} = SPECTRA-Net achieves +19.93\% accuracy on Chameleon with balanced performance (vs. SOTA: 95\% real, 26\% fake). On recent generators (FLUX, DALL-E 3, SDXL): 92.11\% ACC, 97.48\% AUC.

WildRF dataset results comparing SPECTRA-Net against SOTA baselines are detailed in Table~\ref{tab:wildrf_alt}.
\begin{table}[!htbp]
\centering
\scriptsize
\caption{Performance on WildRF Dataset.}
\label{tab:wildrf_alt}
\begin{tabular}{@{}lccccc@{}}
\hline
\textbf{Method} & \textbf{ACC} & \textbf{AUC} & \textbf{R-Acc} & \textbf{F-Acc} & \textbf{mAP} \\
 & \textbf{(\%)} & & \textbf{(\%)} & \textbf{(\%)} & \textbf{(\%)} \\
\hline
SOTA (trained ProGAN) & 68.3 & -- & -- & -- & 79.9 \\
SOTA (trained WildRF) & 85.7 & -- & -- & -- & 93.7 \\
\textbf{SPECTRA-Net (zero-shot)} & \textbf{75.85} & \textbf{0.941} & 52.82 & \textbf{97.78} & \textbf{94.85} \\
\hline
\end{tabular}

\end{table}

\noindent Without any WildRF training, \textbf{SPECTRA-NET} surpasses SOTA trained on ProGAN (+7.55\% ACC, +14.95 mAP) and achieves competitive performance vs. in-domain SOTA (93.7 vs. 94.85 mAP).

RRDataset test split results are shown in Table~\ref{tab:rrdataset_test}, while performance train and validation splits is reported in Table~\ref{tab:rrdataset_trainval}.

\begin{table}[!htbp]
\centering
\scriptsize
\caption{Performance on RRDataset test splits.}
\label{tab:rrdataset_test}
\begin{tabular}{@{}lccccc@{}}
\hline
\textbf{Split} & \textbf{Method} & \textbf{Fake Acc} & \textbf{Real Acc} & \textbf{ACC} & \textbf{AUC} \\
 & & \textbf{(\%)} & \textbf{(\%)} & \textbf{(\%)} & \\
\hline
\multirow{2}{*}{Original (17K)} 
& SOTA & 95.52 & 91.18 & 89.59 & -- \\
& \textbf{SN} & 76.74 & 91.18 & 83.96 & 0.902 \\
\hline
\multirow{2}{*}{Redigitalized (17K)} 
& SOTA & 96.22 & 89.48 & 89.59 & -- \\
& \textbf{SN} & 45.36 & 89.48 & 67.42 & 0.706 \\
\hline
\multirow{2}{*}{Transfer (17K)}
& SOTA & 95.09 & 85.81 & 89.59 & -- \\
& \textbf{SN} & 56.53 & 85.81 & 71.17 & 0.703 \\
\hline
\end{tabular}

\end{table}

\begin{table}[!htbp]
\centering
\caption{Performance on RRDataset train and validation splits.}
\label{tab:rrdataset_trainval}
\scriptsize
\begin{tabular}{@{}lcccc@{}}
\hline
\textbf{Evaluated on} & \textbf{Fake Acc (\%)} & \textbf{Real Acc (\%)} & \textbf{ACC (\%)} & \textbf{AUC} \\
\hline
Train split (2.5K) & 79.76 & 91.04 & 85.40 & 0.919 \\
Val split (1.5K) & \textbf{82.00} & \textbf{92.00} & \textbf{87.0} & \textbf{0.914} \\
\hline
\end{tabular}

\end{table}

\noindent Extreme degradations severely impact fake detection: 76.74\% (original) drops to 45.36\% (redigitalized) for SN, while SOTA maintains 95--96\% but with unknown overall accuracy. Real accuracy remains stable (85--92\%). Train/Val evaluation: 87.0\% ACC on validation split.

\subsection{Ablation Study: JPEG Compression and Gaussian Blur}
\label{sec:ablation_robustness}
Real-world images undergo compression and blur from platform transmission. We conduct progressive fine-tuning on RRDataset with five degradation levels (experiments): \textbf{No degradation} ($\sigma = 0.0$, $Q = 100$), \textbf{Light degradation} ($\sigma = 0.5$, $Q = 90$), \textbf{Moderate degradation} ($\sigma = 1.5$, $Q = 75$), \textbf{Heavy degradation} ($\sigma = 2.5$, $Q = 50$), and \textbf{Extreme degradation} ($\sigma = 4.0$, $Q = 30$), where $\sigma$ is the Gaussian blur radius and $Q$ is the JPEG quality factor.  

Each stage trains 50 epochs (AdamW, lr=$1 \times 10^{-5}$, batch=16) on degraded data, cumulatively building from previous stage weights. This progressive hardening prevents catastrophic forgetting while adapting to real world distortions the tables below present the results.

Table~\ref{tab:ablation_chameleon} evaluates the impact of progressive degradation training on the Chameleon benchmark.

\begin{table}[!htbp]
\centering
\caption{Performance on Chameleon dataset across the five levels}
\label{tab:ablation_chameleon}
\begin{tabular}{@{}lccccc@{}}
\hline
\textbf{Degradation Level} & \textbf{ACC} & \textbf{AUC} & \textbf{R-Acc} & \textbf{F-Acc} \\
 & \textbf{(\%)} & \textbf{(\%)} & \textbf{(\%)} & \textbf{(\%)} \\
\hline
Level 1 & 89.36 & 95.23 & 92.32 & 85.43 \\
Level 2 & 88.08 & 94.20 & 93.68 & 80.63 \\
Level 3 & 88.42 & 94.77 & 92.06 & 83.57 \\
Level 4 & 89.07 & 95.62 & 90.35 & \textbf{87.37} \\
Level 5 & \textbf{90.04} & \textbf{96.18} & \textbf{94.28} & 84.40 \\
\hline
\end{tabular}%

\end{table}

Extreme degradation training achieves the highest AUC (96.18\%) and accuracy (90.04\%), with real detection improving monotonically (92.32\% → 94.28\%) while fake detection remains stable (80.63--87.37\%). Heavy augmentation forces reliance on robust multi-view features over fragile high-frequency artifacts. Table~\ref{tab:ablation_wildrf} shows degradation training impact on WildRF.
\begin{table}[!htbp]
\centering
\caption{Performance on WildRF dataset across  the five levels.}
\label{tab:ablation_wildrf}
\begin{tabular}{@{}lcccccc@{}}
\hline
\textbf{Degradation Level} & \textbf{ACC} & \textbf{AUC} & \textbf{R-Acc} & \textbf{F-Acc} & \textbf{mAP} \\
 & \textbf{(\%)} & \textbf{(\%)} & \textbf{(\%)} & \textbf{(\%)} & \textbf{(\%)} \\
\hline
Level 1 & 86.13 & 93.57 & 88.31 & 84.06 & 94.61 \\
Level 2 & 86.84 & 93.78 & \textbf{90.80} & 83.08 & 94.89 \\
Level 3 & \textbf{86.96} & \textbf{94.52} & 84.21 & \textbf{89.57} & \textbf{95.54} \\
Level 4 & 85.93 & 94.60 & 79.92 & 91.65 & 95.42 \\
Level 5 & 86.26 & 93.54 & 90.00 & 82.71 & 94.68 \\
\hline
\end{tabular}%

\end{table}

\noindent Moderate degradation (Level 3) achieves optimal performance (mAP 95.54\%, AUC 94.52\%). Unlike Chameleon, extreme augmentation proves counterproductive on WildRF, as the dataset's inherent real-world variability provides sufficient regularization without additional aggressive degradations. Table~\ref{tab:ablation_balanced} shows progressive degradation impact on the Balanced dataset.

\begin{table}[!htbp]
\centering
\caption{Performance on Balanced Dataset across the five levels.}
\label{tab:ablation_balanced}

\begin{tabular}{lccccc}
\hline
\textbf{Degradation Level} & \textbf{ACC} & \textbf{AUC} & \textbf{Real Acc} & \textbf{Fake Acc} \\
 & (\%) & (\%) & (\%) & (\%) \\
\hline
Level 1 & 92.03 & 97.27 & \textbf{96.18} & 87.87 \\
Level 2 & 92.78 & 97.10 & 94.74 & 90.82 \\
Level 3 & 91.04 & 97.69 & 86.65 & 95.43 \\
Level 4 & 90.49 & 98.10 & 84.34 & \textbf{96.65} \\
Level 5 & \textbf{94.33} & \textbf{98.22} & 95.85 & 92.82 \\
\hline
\end{tabular}

\end{table}

\noindent Extreme degradation (Level 5) achieves superior performance (94.33\% ACC, 98.22\% AUC) with balanced class accuracy (95.85\% real, 92.82\% fake). Levels 3-4 favor fake detection (95.43--96.65\%) at the cost of real accuracy, while extreme augmentation optimally balances both classes.

The tables below evaluate the impact of the progressive degradation training on the 4 splits of the RRdataset

Table~\ref{tab:ablation_rrdataset_val} presents the Validation split results while table~\ref{tab:ablation_rrdataset_original} presents the test split results.

\begin{table}[!htbp]
\centering
\caption{Performance on RR validation split across the five levels.}
\label{tab:ablation_rrdataset_val}

\begin{tabular}{@{}lcccc@{}}
\hline
\textbf{Degradation} & \textbf{ACC (\%)} & \textbf{AUC (\%)} & \textbf{R-Acc (\%)} & \textbf{F-Acc (\%)} \\
\hline
Level 1 & 93.60 & 97.97 & \textbf{94.40} & 92.80 \\
Level 2 & \textbf{94.00} & \textbf{98.80} & 90.80 & \textbf{97.20} \\
Level 3 & 92.40 & 98.86 & 86.40 & 98.40 \\
Level 4 & 90.00 & 98.66 & 82.00 & 98.00 \\
Level 5 & 92.00 & 98.52 & 86.40 & 97.60 \\
\hline
\end{tabular}%

\end{table}

\begin{table}[!htbp]
\centering
\caption{Performance on RR original split across the five levels.}
\label{tab:ablation_rrdataset_original}

\begin{tabular}{@{}lcccc@{}}
\hline
\textbf{Degradation} & \textbf{ACC (\%)} & \textbf{AUC (\%)} & \textbf{R-Acc (\%)} & \textbf{F-Acc (\%)} \\
\hline
Level 1 & 92.21 & 97.82 & \textbf{93.55} & 90.87 \\
Level 2 & \textbf{93.24} & \textbf{98.58} & 90.39 & \textbf{96.08} \\
Level 3 & 90.38 & 98.56 & 82.88 & 97.88 \\
Level 4 & 88.26 & 98.28 & 78.40 & 98.13 \\
Level 5 & 89.05 & 98.04 & 81.01 & 97.08 \\
\hline
\end{tabular}%

\end{table}

\noindent Clean splits (Validation, Original) achieve optimal performance at Level 2 with 94.00\% and 93.24\% ACC respectively, demonstrating that light degradation ($\sigma$=0.5, Q=90) provides sufficient regularization. Beyond Level 2, accuracy degrades as excessive augmentation introduces artifacts that hinder detection on pristine test data, with Level 4 showing 4--5\% ACC drops.

Table~\ref{tab:ablation_rrdataset_redigital} reports performance on the Redigitalized split, while Table~\ref{tab:ablation_rrdataset_transfer} reports performance on the Transfer split.

\begin{table}[!htbp]
\centering
\caption{Performance on RR redigitalized split across the five levels.}
\label{tab:ablation_rrdataset_redigital}

\begin{tabular}{@{}lcccc@{}}
\hline
\textbf{Degradation} & \textbf{ACC (\%)} & \textbf{AUC (\%)} & \textbf{R-Acc (\%)} & \textbf{F-Acc (\%)} \\
\hline
Level 1 & 73.26 & 85.97 & \textbf{96.16} & 50.35 \\
Level 2 & 78.72 & 87.47 & 90.02 & 67.41 \\
Level 3 & 83.33 & 91.53 & 83.63 & 83.02 \\
Level 4 & 84.41 & 92.62 & 82.14 & \textbf{86.68} \\
Level 5 & \textbf{84.85} & \textbf{92.99} & 86.76 & 82.94 \\
\hline
\end{tabular}%

\end{table}

\begin{table}[!htbp]
\centering
\caption{Performance on RR transfer split across the five levels.}
\label{tab:ablation_rrdataset_transfer}

\begin{tabular}{@{}lcccc@{}}
\hline
\textbf{Degradation} & \textbf{ACC (\%)} & \textbf{AUC (\%)} & \textbf{R-Acc (\%)} & \textbf{F-Acc (\%)} \\
\hline
Level 1 & 69.59 & 89.39 & \textbf{99.11} & 40.08 \\
Level 2 & 80.21 & 93.86 & 97.48 & 62.94 \\
Level 3 & 89.72 & 96.69 & 94.31 & 85.13 \\
Level 4 & \textbf{90.61} & \textbf{97.09} & 93.32 & \textbf{87.89} \\
Level 5 & 89.14 & 96.76 & 95.07 & 83.21 \\
\hline
\end{tabular}%

\end{table}

\noindent The degraded splits require aggressive augmentation. Fake detection collapses at Level 1 (40.08\%, 50.35\% F-Acc) but recovers with progressive hardening: Redigitalized reaches 84.85\% ACC at Level 5 (+11.59\%), Transfer peaks at Level 4 (90.61\%, +21.02\%). Training on extreme degradations ($\sigma$=2.5--4.0, Q=30--50) is essential for real-world robustness.

\section{Conclusion}

This work introduces SPECTRA-Net, a scalable, explainable pipeline for AI-generated image detection under realistic, in-the-wild conditions. By leveraging multi-view tensor representations combining semantic, frequency-domain, statistical, and spatial cues, SPECTRA-Net achieves robust detection across diverse benchmarks (Chameleon, WildRF, AI-GenBench, bm-real) and demonstrates substantial robustness gains through finetuning on RRDataset. Our systematic ablation studies on JPEG compression and Gaussian blur reveal the critical importance of training on degraded data and provide actionable insights for forensic practitioners. The introduction of patch-wise explainability enhances interpretability for high-stakes forensic and legal applications. Future work will explore adaptation to emerging generative models and integration with content verification workflows on social media platforms.

{\small
\bibliographystyle{plainnat}
\bibliography{references}
}

\end{document}